# The UN Parallel Corpus Annotated for Translation Direction

Elad Tolochinsky, Ohad Mosafi, Ella Rabinovich, Shuly Wintner

September 18, 2017

**Abstract**

This work distinguishes between translated and original text in the UN protocol corpus. By modeling the problem as classification problem, we can achieve up to 95% classification accuracy. We begin by deriving a parallel corpus for different language-pairs annotated for translation direction, and then classify the data by using various feature extraction methods. We compare the different methods as well as the ability to distinguish between translated and original texts in the different languages. The annotated corpus is publicly available.

## 1 Introduction

This work aims to distinguish between original texts and translated texts. Given a document written in some language, we wish to accurately say whether the document was originally written in that language or it was translated to it. To distinguish such texts, we use a linguistic concept called translationese. Translated texts, in any language, can be considered a dialect of this language. We call this dialect translationese. Linguistic researches have proposed several universal properties of translationese which we use in conjunction with supervised machine learning techniques to distinguish between translated texts and non-translated (original) texts. Previous work [2], on this subject have tried to gauge which properties of translationese can be utilized to successfully classify original and translated texts by performing an extensive exploration of that ability of different feature sets. Another work [1], presents a high In-domain classification accuracy using these feature sets on Europal and three additional sub-corpora - Hansard, Literature and TED. However, the cross-domain classification accuracy on Europal, Hansard and Literature corpora, when training on one corpus and testing on another yields a success rate score which slightly outperform chance level. Our work reproduces the results achieved by [2] on a different corpus - the UN parallel corpus. The rest part of this work is an automated derivation of the corpus. We derived five bilingual parallel corpora, from English to any other official UN language (French, Spanish, Russian, Arabic and Chinese).

## 2 Derivation of the Corpora

The base of this work is the UN parallel corpus which is described at length in [3]. The corpus is structured in a directory hierarchy, each language has a directory which holds the documents in that language. The documents are stored in a directory tree inside the appropriate language folder in a way that the relative path of a document inside a specific language directory is the same for all language directories, for example the file add_1.xml has a French version at the path \fr\1990\trans\wp_29\1999\14\add_1.xml and an English version at the path \en\1990\trans\wp_29\1999\14\add_1.xml. Every language pair has an additional directory which contains link files. The link files defines the translation direction of two documents and they reside at the same relative path as the documents, thus the link file of add_1.xml is located at \fr_en\1990\trans\wp_29\1999\14\add_1.lnk .

To derive a bilingual parallel corpus, we pick a language and traverse its directory alongside the English directory (Note that we must start the derivation from the non-English language, due to the fact that the link file matches non-English sentences to their English translation but not the opposite). For each sentence, we use the link file to determine if the sentence is in original or translated. The outputs of the process are 3 files for each language: a file containing English sentences, a file in which each line is the translation of the parallel line in the English file and a third file which species the original language of each line. While processing the documents, many documents were filtered out of the corpus due to distinct reasons, such as: documents that had no corresponding document in English, documents that did not specify their source language (either in English or in the source language), documents which source language did not match the current language, etc. After obtaining the valid documents we filtered out invalid sentences which include sentences which language tag was different then the language tag of the file and sentences which has no destination at the link file.

|  | French-English | Spanish-English | Russian-English | Arabic-English | Chinese-English |
| --- | --- | --- | --- | --- | --- |
| Initial number of protocols | 185,800 | 130,275 | 138,168 | 116,751 | 95,946 |
| Number of valid protocols in every language and percentage (from the initial amount) | 4,567 / 51,807 (2.8% / 32.6%) | 1,957 / 36,603 (1.5% / 28%) | 706 / 25,218 (0.5% / 18%) | 1,690 / 29,053 (1.4% / 24.8%) | 10 / 8,290 (0.01% / 8.6%) |
| Number of valid sentences | 8,926,298 | 6,638,552 | 3,740,232 | 4,175,839 | 1,689,598 |
| Number of valid sentences in every language and their percentage | 773,276 / 8,153,022 (6.5% / 68.8%) | 447,445 / 6,191,107 (5% / 70%) | 107,737 / 3,632,495 (2% / 73%) | 88,263 / 4,087,576 (1.4% / 68.5) | 4,768 / 1,684,830 (0.2% / 73%) |

Table 1: Corpora derivation results

Results of the derivation are summarized in Table 1.

## 3 Classification

The methods we employ in this work are machine learning, more specially classification. Generally, given a set of labeled vectors X * Y where X belongs to Rd and Y = {0, 1} which are drawn from some distribution D. A classification function is a function f: Rd → Y such that with high probability f (x) = y for every (x; y) belongs to D. The process of computing such a function is called training and is based upon feeding the learning algorithm with known examples from which it can learn. In this work we employ well know learning algorithms such as SVM and logistic regression. We will use the implementation of Python's sklearn package. To sum up, to distinguish between original texts and translations, we transform a document to a vector, we label that vector according to the class of the document (translated or original) and we then proceed to train the appropriate machine until we obtain a function which distinguishes between translated and original.

## 4 From Documents to Vectors

In the previous section we described a method in which we can distinguish between different classes of vectors. We are left with the problem of representing a document by a multi-dimensional numeric vector. There are many ways to represent a document as a vector, perhaps the simplest is called 'bag-of-words' in which every document is represented by a vector of counters, every entry in the vector represent the number of occurrences of a corpus word in this specific document. However, such simple representation may not be helpful in distinguishing between original and translated texts. That is where we employ translationese. We use the hypothesized universal properties of translationese to derive a numeric representation of documents. As these properties represent the dialect of translated properties, we intuitively expect that will produce accurate classification results. The work in [2] have compared many of the universal properties of the translationese and have found the ones that are most effective for distinguishing original from translated.

We used the following properties:

Function word, POS trigrams, POS bigrams, POS trigrams and function words.

In order to obtain a dataset for training our learning algorithms we tokenize the text files and add part of speech tagging, we then break up a file to chunks of about 2000 tokens. Each chunk will be transformed into a vector according to the chosen property.

Function words -- each chunk is transformed to a vector, where each entry in the vector represents the frequency of a function word in the chunk. We then normalize this quantity by multiplying it by n=2000 where n is the size of the chunk.

|  |  | FW | Trigrams | Bigrams |
|---|---|---|---|---|
| Each chunk contains lines which belong to one language out of French, Spanish, Russian and Arabic. | All languages (excluding Chinese) | 83% | 85% | 86.65% |
|  | French-English | 86.5% | 87% | 88.28% |
|  | Spanish-English | 86% | 86.8% | 88.5% |
|  | Russian-English | 91.38% | 92.48% | 92.86% |
|  | Arabic-English | 89.71% | 93.6% | 94.5% |
| Lines in chunks are randomly distributed across languages | All languages | 90.53 | 92.05% | 93.21% |

Table 2: Classification results

POS trigrams -- each chunk is transformed to a vector, where each entry in the vector is the number of occurrences of a POS trigram in the chunk from a list of 400 top trigrams.

POS bigrams -- each chunk is transformed to a vector, where each entry in the vector is the number of occurrences of a POS bigram in the chunk from a list of 400 top bigrams.

5   **Experiments and Results**

After obtaining the bilingual parallel corpora we divided the sentences at the English file of the corpus to tokens and added POS tags. The tokenizer we used is NLTK's tweet tokenizer and POS tagging was done with OpenNLP. After obtaining tokenized and tagged documents we start dividing each document to chunks of about 2000 tokens. Each chunk is comprised from sentences that are either all original or all translated.

We then transform each chunk to a vector using one of the methods described above and label the vector accordingly. From this process we obtain a set of vectors that was derived from each language. We merge all these vectors and feed them to a classier. Afterwards we classify each language separately to see in which languages the distinction from original to speech is easier. Finally, we reproduced the chunks, only this time we shuffled together translated English sentences from all of the languages and sentences originally in English were shuffled together from different bilingual corpora as well. Then divided them to chunks. The results are depicted in Table 2. In all tests we omitted the samples from the Chinese corpus since it was too small. All data sets were balanced, so the baseline is 50%.

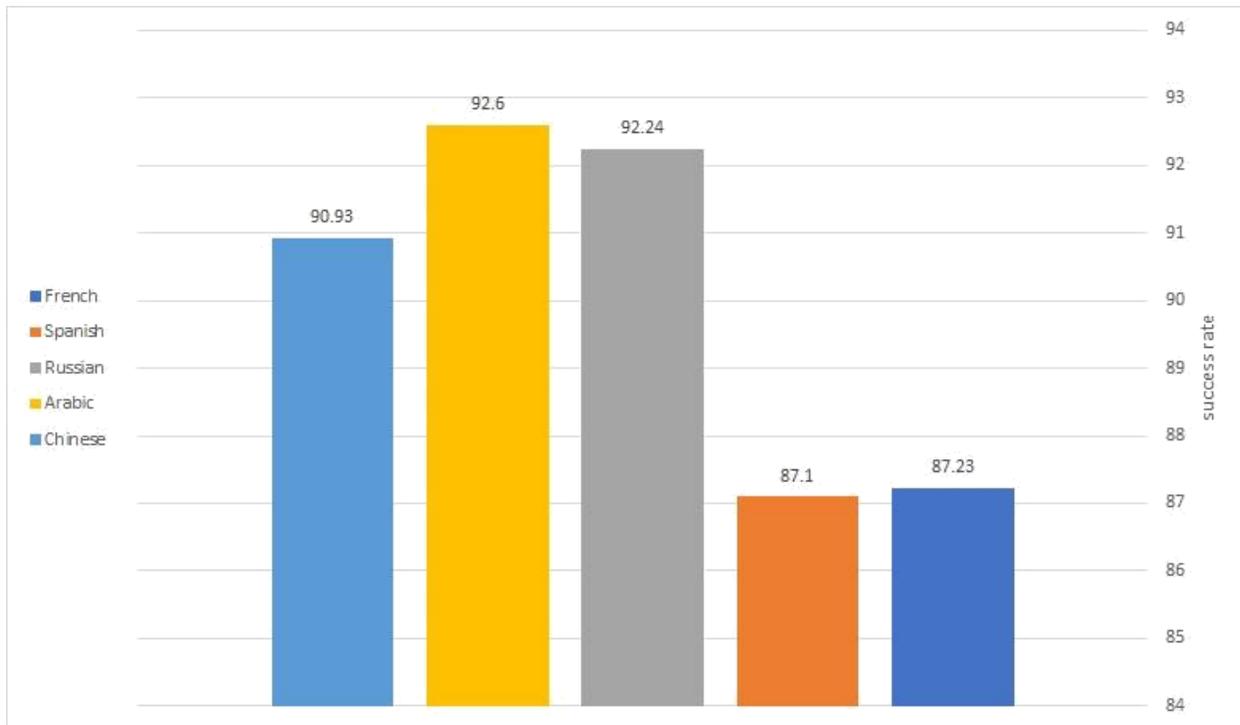

Figure 1: Comparing classification results between languages (the average of the three feature extraction options).

The comparison between the classification results is depicted in Figure 1.

We also tested the effects of changing the number of tokens in a chunk, where lines in chunks are randomly distributed from different languages. We produced various datasets from different sized chunks. Each dataset was balanced by taking all the translated chunks and randomly choosing the same number of original chunks before classification. The results are depicted in Figure 2. The number of chunks per chunk size can be seen at

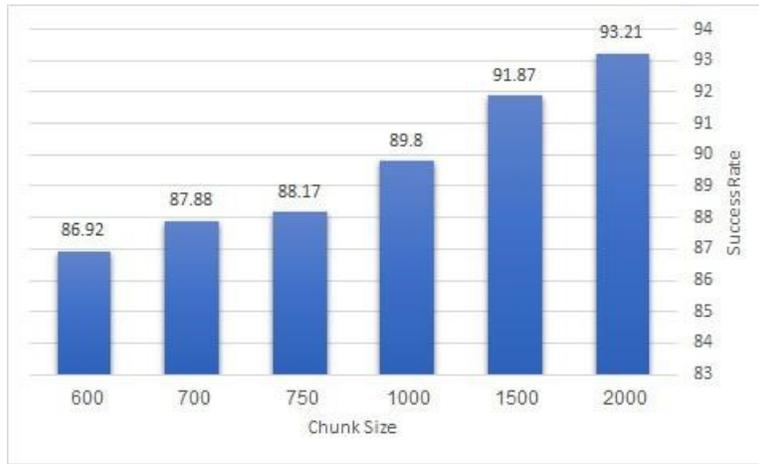

Figure 2: Classification results for various amounts of tokens in a chunk. Tests were performed using bigrams as features

| Chunk Size | Number of samples |
|---|---|
| 2000 | 36,632 |
| 1500 | 48,422 |
| 1000 | 71,412 |
| 750 | 93,664 |
| 700 | 99,878 |
| 600 | 115,210 |

Table 3: Number of samples per chunk size

Finally, we experimented with combining the features, we transferred each chunk to a vector by using function words and POS trigrams. We conducted several experiments, each with different number of POS. In all the experiments we used randomly distributed chunks with 2000 tokens taken from all the bilingual corpora. The results are shown in Figure 3.

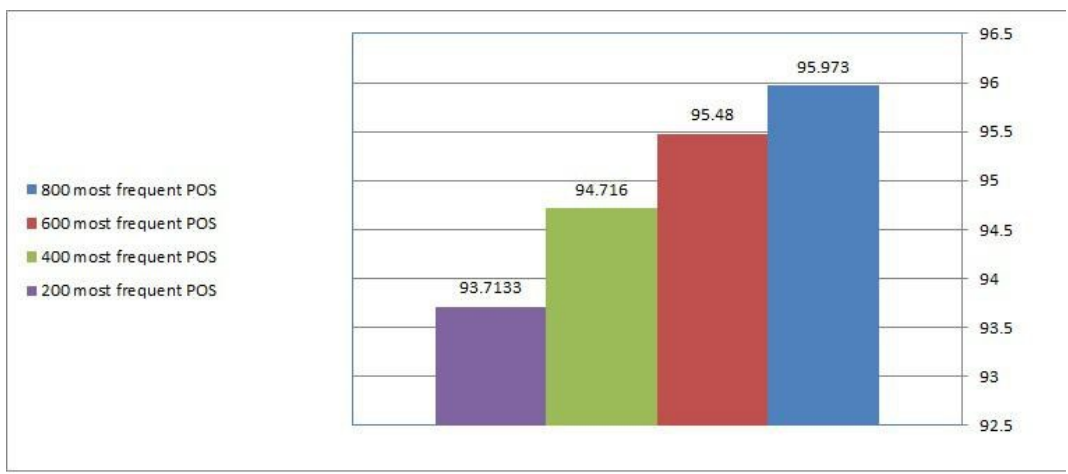

(a) Using trigrams and function words

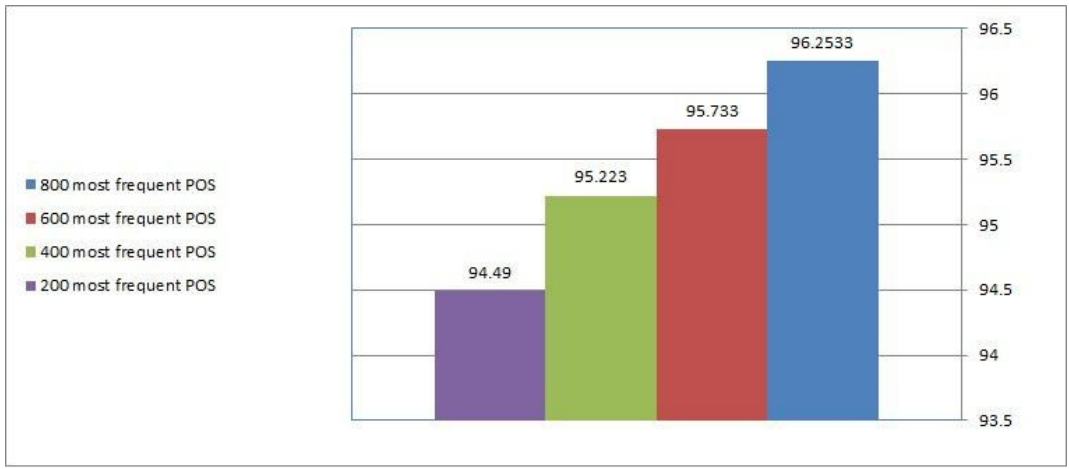

(b) Using bigrams and function words

Figure 3: Success rate of classification using function words and POS for varying amount of top POS to use for classification.

## 6   Conclusions

The results of our work raise several conclusions: first, the corpora we derived are valid, for if they were not so we would not have been able to classify with high accuracy. Furthermore, we can see that the different languages are classified more easily than others, this is somewhat intuitive, since French is much more similar to English then Russian. Last, we see that building chunks out of sentences that were randomly chosen across all languages yield much better results, this is reasonable, since choosing sentences from all languages negates noises (personal speech style, subject, language of origin, etc.) and we are left with a one-dimensional data - translate or origin.

The corpus is available per request from the authors - shuly@cs.haifa.ac.il, ellarabi@gmail.com.